\documentclass[sigconf, nonacm]{acmart}

\usepackage[T1]{fontenc}
\usepackage[utf8]{inputenc}
\usepackage{amsmath}
\usepackage{graphicx}
\usepackage{subcaption}
\usepackage{caption}
\usepackage{xspace}

\newcommand{\stab}{\vspace{1.2ex}\noindent}
\newcommand{\stitle}[1]{\stab\noindent{\textbf{#1}}}

\newcommand{\eg}{\textit{e.g.,}\xspace}
\newcommand{\model}{TOFFEE\xspace}

\newcommand\vldbdoi{10.14778/3827998.3828110}
\newcommand\vldbpages{4738 - 4741}
\newcommand\vldbvolume{19}
\newcommand\vldbissue{12}
\newcommand\vldbyear{2026}
\newcommand\vldbauthors{\authors}
\newcommand\vldbtitle{\shorttitle}
\newcommand\vldbavailabilityurl{https://github.com/wang0702/toffee}
\newcommand\vldbpagestyle{empty}

\begin{document}
\title{Demonstrating TOFFEE: A Learned System for Synthesizing\\ Data Agent Trajectories at Scale}

\author{Ziting Wang}
\affiliation{%
  \institution{Nanyang Technological University}
  \country{Singapore}
}
\email{ziting001@e.ntu.edu.sg}

\author{Yin Li}
\affiliation{%
  \institution{Nanyang Technological University}
  \country{Singapore}
}
\email{yin010@e.ntu.edu.sg}

\author{Zuhao Yang}
\affiliation{%
  \institution{Nanyang Technological University}
  \country{Singapore}
}
\email{YANG0756@e.ntu.edu.sg}

\author{Xiuchang Li}
\affiliation{%
  \institution{Huawei}
  \country{China}
}
\email{lixiuchang@huawei.com}

\author{Jiale Bai}
\affiliation{%
  \institution{Industrial and Commercial Bank of China Limited, China}
}
\email{baijl2@sdc.icbc.com}

\author{Gao Cong}
\affiliation{%
  \institution{Nanyang Technological University}
  \country{Singapore}
}
\email{gaocong@ntu.edu.sg}

\begin{abstract}
LLM-powered data agents are playing an increasingly important role in data-driven decision making. However, existing data agents struggle to generalize to unseen data environments and analytical workflows, 
especially in heterogeneous enterprise settings.
This creates 
a growing need 
for synthesizing high-quality data agent trajectories that capture complex analytical workflows for given data environments. Such trajectories 
support two key downstream uses: they 
can
serve as supervised finetuning (SFT) data that adapts data agent models to the target domain, and as in-context learning (ICL) demonstrations
to guide general-purpose LLMs 
in unfamiliar data environments. Thus, we introduce \model, a 
system for synthesizing high-quality data agent trajectories from given data environments via Monte Carlo Tree Search (MCTS) with adaptive model selection and cross-task prefix reuse. We show that \model can effectively 
generate scalable trajectory data
for complex analytical tasks across heterogeneous
environments. In this demonstration, we present the system framework of \model, including its task pool construction, trajectory explorer, and learned cost model. We also introduce the web interface of \model and its workflow, and 
demonstrate two end-to-end scenarios:
trajectory synthesis for data agent finetuning, and demonstration-augmented data agent reasoning.
\end{abstract}

\maketitle

\pagestyle{\vldbpagestyle}
\begingroup\small\noindent\raggedright\textbf{PVLDB Reference Format:}\\
\vldbauthors. \vldbtitle. PVLDB, \vldbvolume(\vldbissue): \vldbpages, \vldbyear.\\
\href{https://doi.org/\vldbdoi}{doi:\vldbdoi}
\endgroup
\begingroup
\renewcommand\thefootnote{}\footnote{\noindent
This work is licensed under the Creative Commons BY-NC-ND 4.0 International License. Visit \url{https://creativecommons.org/licenses/by-nc-nd/4.0/} to view a copy of this license. For any use beyond those covered by this license, obtain permission by emailing \href{mailto:info@vldb.org}{info@vldb.org}. Copyright is held by the owner/author(s). Publication rights licensed to the VLDB Endowment. \\
\raggedright Proceedings of the VLDB Endowment, Vol. \vldbvolume, No. \vldbissue\ %
ISSN 2150-8097. \\
\href{https://doi.org/\vldbdoi}{doi:\vldbdoi} \\
}\addtocounter{footnote}{-1}\endgroup

\ifdefempty{\vldbavailabilityurl}{}{
\vspace{.3cm}
\begingroup\small\noindent\raggedright\textbf{PVLDB Artifact Availability:}\\
The source code, data, and/or other artifacts have been made available at \url{\vldbavailabilityurl}.
\endgroup
}

\section{Introduction}
\label{sec:intro}

\begin{figure}[t!]
  \centering
  \includegraphics[width=0.90\columnwidth]{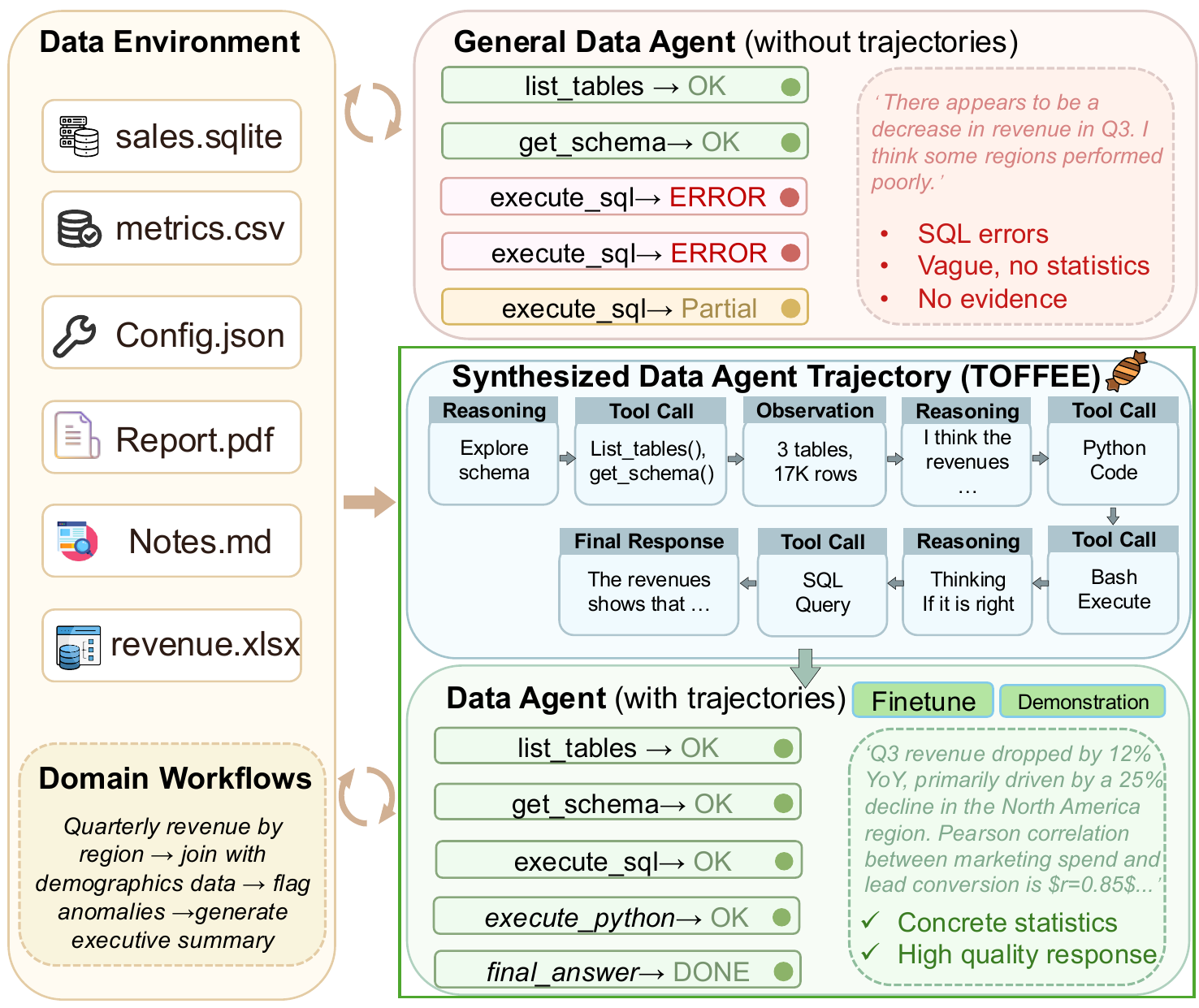}
  \caption{Data agent trajectory synthesis. Top: without trajectories, the agent produces SQL errors and vague conclusions. Middle: synthesized trajectories. Bottom: trajectory-augmented agent via SFT or ICL.}
  \label{fig:example}
\end{figure}

Data agents that analyze data by interleaving reasoning with tool execution (\eg SQL, Python)
have attracted growing attention for data-driven decision-making~\cite{jing2025dsbench, lai2025kramabench}.
Correspondingly, major data lake and data warehouse platforms are beginning to integrate such capabilities, \eg Databricks Genie, Snowflake Cortex Analyst, and BigQuery data agents.
However, existing data agents struggle to generalize to unseen data environments and analytical workflows~\cite{jing2025dsbench},
especially in heterogeneous
enterprise environments.
Synthesizing high-quality data agent trajectories, i.e., multi-step sequences of reasoning, tool invocations, and execution results, for a given data environment can bridge this gap.
Figure~\ref{fig:example} illustrates this.
The top row shows a data agent operating without 
trajectories: it 
makes repeated SQL errors and returns a vague conclusion. The middle row shows synthesized trajectories 
generated by running agent actions against the actual data environment.
When used for SFT or ICL (as shown in the bottom row), 
these trajectories help the same agent avoid errors and produce concrete evidence
(\eg Q3 revenue dropped 12\% YoY, Pearson $r{=}0.85$).
However, acquiring such trajectories at scale remains difficult~\cite{li2025omnisql, lin2025lead}. Single-pass generation discards all progress upon any step failure, while best-of-$N$ sampling incurs redundant computation across independent attempts.
Building such a system faces the following challenges.

\begin{figure*}[t!]
  \centering
  \includegraphics[width=0.80\textwidth]{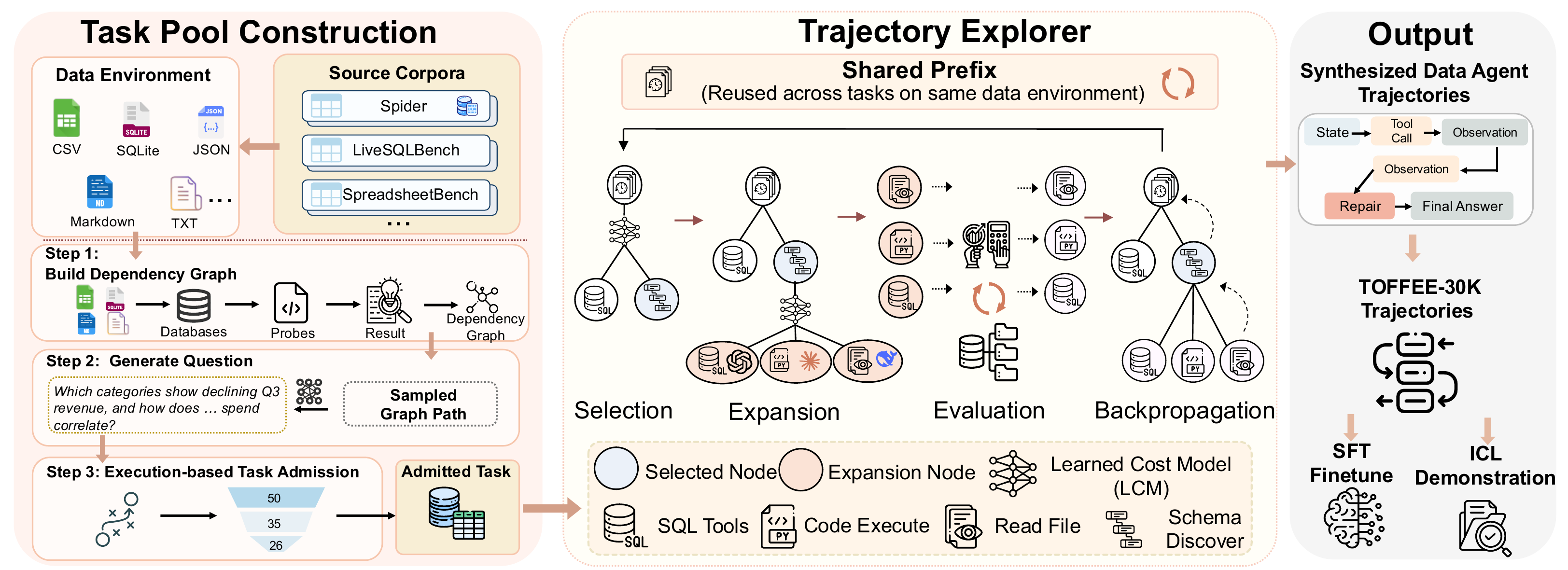}
  \caption{System overview of \model: task pool construction (left), trajectory explorer (center), and export of data agent trajectories for SFT or Demonstration ICL (right).}
  \label{fig:toffee-overview}
\end{figure*}

\stitle{Challenges.}
First, data agent trajectory synthesis requires a large pool of diverse analytical tasks, \eg ``compute quarterly revenue by region'' or ``correlate marketing spend with lead conversion rates,'' each grounded in the tables and files of the target data environment.
Prompting LLMs to generate tasks from schemas often produces trivial or unsolvable questions, and manually curating tasks does not scale.
The first challenge is \emph{how to automatically construct realistic, solvable tasks from
user-specified data environments} (\textbf{C1}).
Second, synthesizing data agent trajectories is inherently complex: each step requires a full cycle of LLM inference and tool execution and can choose among tool actions (\eg SQL query, Python script, or schema discovery) generated by different LLMs, so the number of possible trajectories grows exponentially with the number of steps.
The second challenge is \emph{how to synthesize high-quality trajectories in such a vast search space} (\textbf{C2}).
Third, each synthesis step requires choosing an inference configuration: which LLM to invoke, how many prior steps to feed as prompt context, and whether to enable extended thinking of LLMs.
For example, a smaller model can list tables and read column names, but fails when the step requires writing a multi-table join query; a larger model handles the join correctly but wastes budget on such routine steps.
The third challenge is \emph{how to select the right inference configuration at each step based on the step's difficulty} (\textbf{C3}).

\stitle{Our Method.}
To address these challenges, we introduce \model, a budget-aware trajectory synthesis system.
First, to automatically construct a large pool of analytical tasks, we design a \emph{Task Pool Construction} pipeline. It discovers dependencies in the data environment through lightweight scripts, then reverses each dependency into a multi-step analytical task, and verifies that each task is both solvable and meaningful (addressing C1).
Second, for a task, to efficiently navigate the vast search space
of trajectory synthesis, we design a \emph{Trajectory Explorer}. It implements MCTS-based search, where every candidate step is executed against the 
data environment, naturally capturing diverse execution paths (\eg error recoveries) (addressing C2).
Third, to select the right inference configuration (\eg which LLM to call) at each step based on the step's difficulty, we design a \emph{Learned Cost Model} (LCM).
The LCM observes execution rewards and learns state-dependent policies
through online reward learning, 
 allowing it to distinguish
 routine steps from complex analytical steps without manual tuning
 (addressing C3).

Preliminary results support this design. Under the same per-task budget, \model synthesizes higher-quality trajectories than single-pass generation and best-of-$N$ sampling. Models finetuned on the synthesized trajectories achieve substantial gains over their base models on KramaBench and DSBench (Section~\ref{sec:results}).

In this demonstration, \model's interactive web interface lets users connect heterogeneous data environments, synthesize trajectories with real-time search tree visualization, and export results for SFT and ICL.
We present two end-to-end scenarios.

\section{Overview of \model}
\label{sec:overview}

The architecture of \model is presented in Figure~\ref{fig:toffee-overview}.
\model takes as input a data environment $\mathcal{E}$, which may contain databases, files, and resources accessible through sandboxed code.
It produces data agent trajectories in three stages: task pool construction (left), trajectory exploration (center), and export for SFT or ICL (right).

\stitle{Task Pool Construction.}
Data agent trajectory search is useful only when tasks are realistic, solvable, and tied to the data environment.
A naive way is to directly prompt the schema into an LLM, but due to this method's limited context of the database, it often yields trivial or unsolvable questions.
To address this, \model constructs a data agent task pool before search, grounding each task in the data environment itself.
As shown on the left of Figure~\ref{fig:toffee-overview}, \model builds tasks from the given data environment.
The environment is either connected by the user or sampled from existing benchmarks such as Spider, LiveSQLBench, and SpreadsheetBench.
It first profiles the environment to collect schema metadata and sample rows, then runs predefined SQL templates to discover data dependencies such as a shared customer ID linking several tables, or order cancellation rates that vary sharply across regions, and finally prompts an LLM to phrase each discovered dependency as a meaningful question for a data agent to answer.
Each dependency is executed before its question is written, and the final result is recorded as the task's answer key.
\model then verifies each candidate task: it replays the solution steps the way an agent would, reruns each query to confirm the results are stable, and rejects questions that an LLM can answer without accessing the data.
It admits only tasks with a deterministic answer and verified execution evidence, ensuring strict isolation from downstream benchmarks to prevent evaluation contamination.

\stitle{Trajectory Explorer.}
The \emph{Trajectory Explorer} synthesizes trajectories for each task through a tree search: each node in the search tree is an analysis state, and each edge is a tool invocation, such as running a SQL query or a Python script, generated by a specific LLM configuration.
Rather than committing to a single path, the Explorer expands multiple branches and keeps the most promising.
The search starts from a shared prefix reused across tasks in the same data environment, so common operations such as schema discovery and data profiling run only once.
At each expansion, the Explorer invokes the LCM, described next, to choose the branching width and inference configuration under the remaining budget.
When a step hits an execution error, the Explorer does not discard the partial trajectory; it branches into a repair path, and the resulting error-diagnosis-repair sequence is retained as training data that teaches downstream agents to recover from realistic mistakes.
A trajectory is stored once a deterministic verifier confirms that its executed outputs reproduce the evidence recorded at task construction and its final answer matches the execution-verified answer key.

\stitle{Learned Cost Model.}
The \emph{Learned Cost Model} (LCM) is a budget-aware controller for the Explorer that selects the LLM and inference configuration for each expansion from the current analysis state and the remaining per-task API budget.
For instance, it routes routine schema discovery to a small, fast model while escalating complex multi-table reasoning to a more capable one.
As shown at the bottom of Figure~\ref{fig:toffee-overview}, at each expansion the LCM extracts a state feature vector encoding trajectory progress, schema coverage, and recent error history, then 
scores each feasible configuration, including the operator, model, context length, and reasoning effort, by combining the predicted reward $\mu$ with an exploration bonus $\sigma$ for less-explored configurations.
It sets the branching width $K$ from the disagreement among top-ranked configurations, narrowing $K$ as the budget shrinks.
The Explorer and LCM form a closed feedback loop: after each step, the system computes a reward from execution success and analytical progress, and updates the LCM's bandit parameters via ridge regression, improving its routing and branching over time.

\stitle{Trajectory Export and Downstream Usage.}
As shown on the right of Figure~\ref{fig:toffee-overview}, each trajectory is a sequence of (state, tool call, observation) tuples, exported in two modes.
For \emph{SFT}, trajectories become multi-turn training samples that teach agent models the full analytical workflow, yielding models such as TOFFEE-9B and TOFFEE-27B.
For \emph{ICL demonstration}, the highest-quality trajectories serve as worked examples: when a user poses a question, the system retrieves a similar trajectory into the LLM prompt, so the model follows the demonstrated workflow on the unfamiliar environment.

\section{Preliminary Experimental Results}
\label{sec:results}

\begin{figure}[t]
  \centering
  \begin{subfigure}[t]{0.48\columnwidth}
    \centering
    \includegraphics[width=\textwidth]{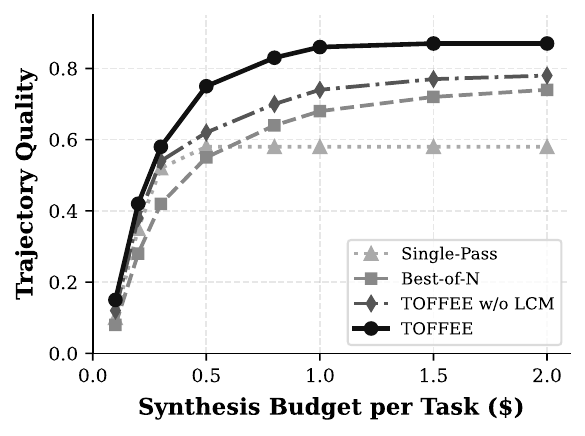}
    \caption{Trajectory quality vs.\ budget.}
    \label{fig:results-a}
  \end{subfigure}
  \hfill
  \begin{subfigure}[t]{0.48\columnwidth}
    \centering
    \includegraphics[width=\textwidth]{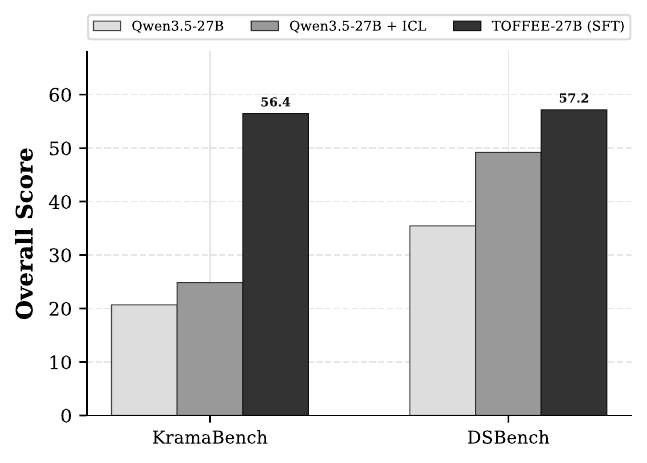}
    \caption{Downstream score.}
    \label{fig:results-b}
  \end{subfigure}
  \caption{Preliminary results on synthesis efficiency and downstream effectiveness.}
  \label{fig:results}
\end{figure}

We conduct experiments on an Ubuntu server, where synthesis workers use API calls to LLMs via OpenRouter.
Trajectory Quality is scored on a 0 to 1 scale from execution signals (\eg step success rate, analytical progress) and a deterministic check of the final answer against each task's execution-verified answer key.
Best-of-$N$ uses $N{=}5$; all methods in Figure~\ref{fig:results-b} share the same agent prompt and tool access.
Downstream evaluation uses KramaBench~\cite{lai2025kramabench} and DSBench~\cite{jing2025dsbench}, with TOFFEE-9B and TOFFEE-27B finetuned from Qwen3.5-9B and Qwen3.5-27B.

Figure~\ref{fig:results-a} compares trajectory quality under varying per-task budgets.
Single-Pass plateaus once the budget covers a single full attempt. Best-of-$N$ improves with more budget but with diminishing returns, as each independent attempt repeats work from scratch.
\model w/o LCM improves over them through MCTS exploration but is limited by uniform configuration at every step.
\model consistently outperforms all baselines, converging to high quality at lower cost, which confirms the value of adaptive per-step model selection via the LCM.
Figure~\ref{fig:results-b} shows the downstream effectiveness of \model trajectories applied to Qwen3.5-27B.
Using \model-synthesized trajectories as ICL demonstrations already yields improvement over the zero-shot baseline.
Finetuning on \model trajectories (TOFFEE-27B) further amplifies the gain, more than doubling the zero-shot score on KramaBench.

\begin{figure*}[t!]
  \centering
  \begin{minipage}[t]{0.43\textwidth}
    \centering
    \includegraphics[width=\textwidth]{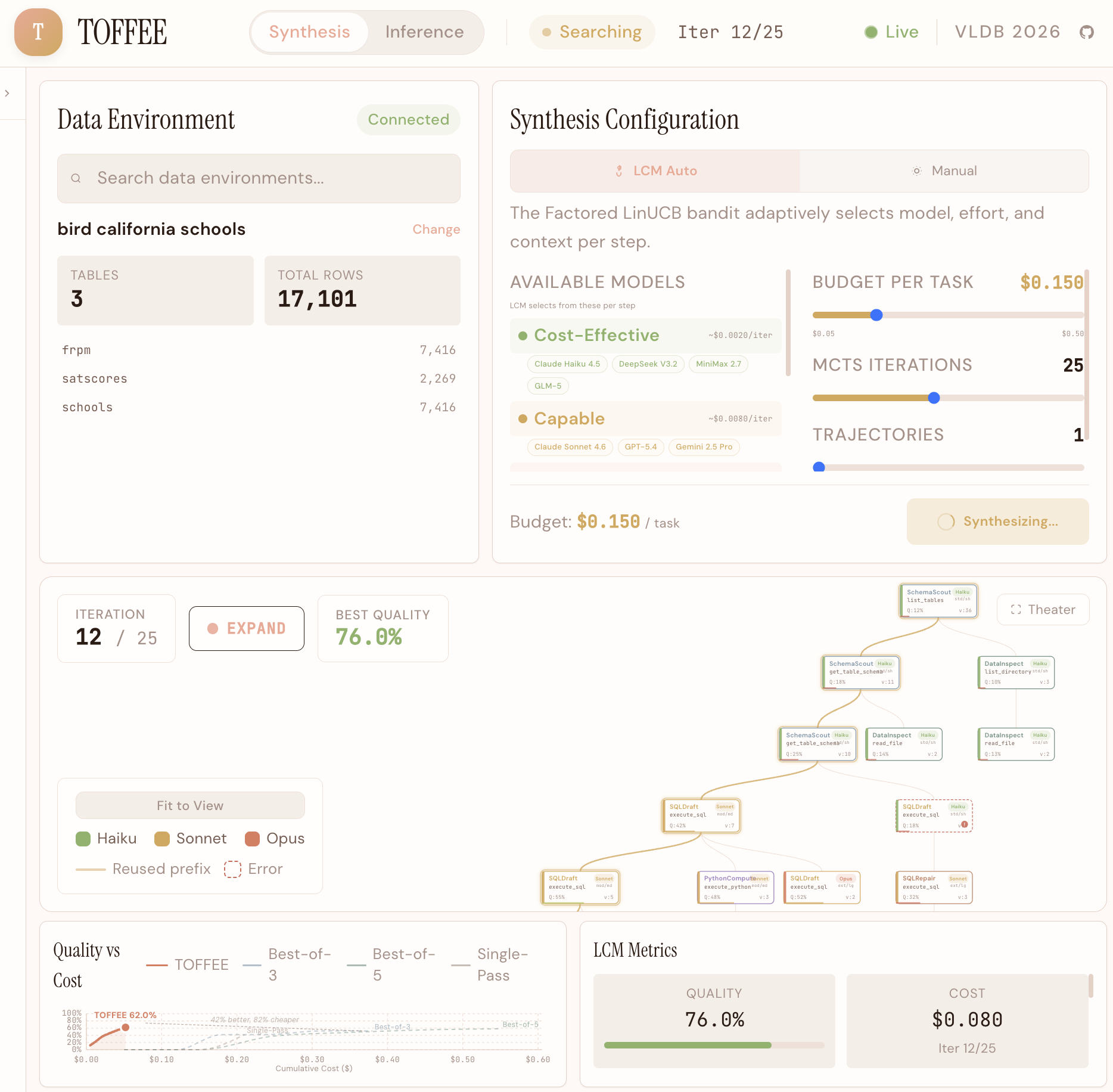}
    \captionof{figure}{Synthesis mode of \model.}
    \label{fig:demo-synthesis}
  \end{minipage}\hfill
  \begin{minipage}[t]{0.43\textwidth}
    \centering
    \includegraphics[width=\textwidth]{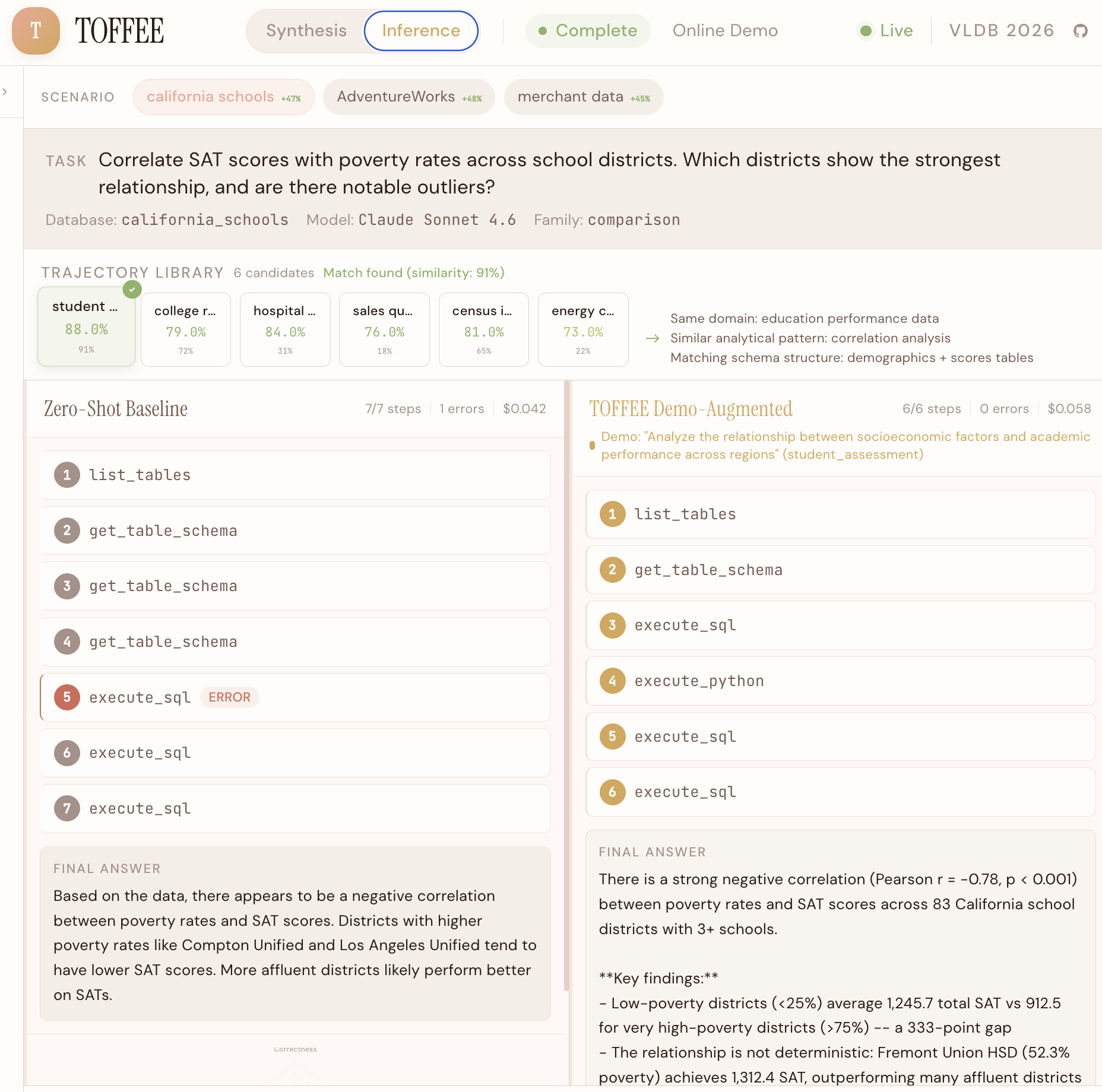}
    \captionof{figure}{Inference mode of \model.}
    \label{fig:demo-inference}
  \end{minipage}
\end{figure*}

\section{Demonstration Scenario}
\label{sec:demo}

The demonstration includes a \emph{Synthesis} mode for trajectory production (Figure~\ref{fig:demo-synthesis}) and an \emph{Inference} mode for demonstration-augmented data agent reasoning (Figure~\ref{fig:demo-inference}).

\subsection{End-to-End Experience}

Figure~\ref{fig:demo-synthesis} shows the Synthesis mode of \model.
Users perform trajectory synthesis in the following four steps.

\noindent \underline{\textit{Step 1: Environment Setup.}}
Users connect a data environment with heterogeneous sources such as databases, CSV files, JSON, and Markdown documents.
The \emph{Data Environment} panel on the top-left of Figure~\ref{fig:demo-synthesis} then lists its table names, row counts, and schema.

\noindent \underline{\textit{Step 2: Synthesis Configuration.}}
In the \emph{Synthesis Configuration} panel on the top-right of Figure~\ref{fig:demo-synthesis}, users assemble a model pool spanning three cost tiers, namely \emph{Cost-Effective}, \emph{Capable}, and \emph{Premium}.
They choose \emph{LCM Auto} mode, where the LCM selects the model, reasoning effort, and context length per step, or \emph{Manual} mode for direct control, and set the per-task budget, MCTS iteration count, and target trajectory count via sliders.

\noindent \underline{\textit{Step 3: Trajectory Synthesis.}}
The MCTS tree visualization in the middle of Figure~\ref{fig:demo-synthesis} renders each node as a card showing the tool type, the color-coded model tier selected by the LCM, and the evaluation score.
For instance, a schema reading node stays in the Cost-Effective tier, while an SQL repair node after an execution error escalates to the Capable or Premium tier, reflecting the LCM's policy of investing more on harder steps.
The \emph{Quality vs Cost} chart tracks quality over iterations against baselines such as Best-of-$N$ and Single-Pass, and the \emph{Metrics} panel shows cost and progress.

\noindent \underline{\textit{Step 4: Export and Downstream Use.}}
After synthesis, users export trajectories as SFT samples or test them as ICL demonstrations in the \emph{Inference} tab (Figure~\ref{fig:demo-inference}).

\subsection{Scenario 1: Trajectory Synthesis for Finetuning}

In Figure~\ref{fig:demo-synthesis}, the user connects a school database with 3 tables and 17K rows, configures a three-tier model pool, and sets the per-task budget and MCTS iteration count.
The system profiles the environment, constructs the task pool, and launches MCTS-based search.
As synthesis proceeds, the tree visualization grows live, with execution-error nodes spawning repair branches and the shared prefix reused across tasks on the same database.
Clicking a node shows the query executed at that step and the result it returned.
The user then exports the synthesized trajectories as SFT training samples for data agent models such as TOFFEE-9B and TOFFEE-27B.
The exported samples follow the standard multi-turn reasoning format and can be loaded directly by common finetuning frameworks.

\subsection{Scenario 2: Demonstration Augmented Data Agent Reasoning}

As shown in Figure~\ref{fig:demo-inference}, in the \emph{Inference} tab, the user selects a scenario and poses an analytical task such as ``Correlate SAT scores with poverty rates across school districts and identify outliers.''

The system maintains a library of 30K trajectories synthesized across hundreds of data environments.
Given the new task, it retrieves the most relevant trajectories ranked by domain and workflow similarity, and injects them as few-shot demonstrations.
Figure~\ref{fig:demo-inference} shows a 91\% similarity match from an education performance database, guiding the agent through a proven workflow.

The interface presents a side-by-side comparison: the zero-shot baseline produces a vague answer with execution errors, while the \model-augmented agent follows a structured workflow, including schema discovery, table joins, and statistical computation, yielding concrete findings (\eg Pearson $r{=}{-}0.78$, $p{<}0.001$) with zero errors.

\begin{acks}
This research is supported by Singapore MOE AcRF Tier-2 grant MOE-T2EP20223-0004.
\end{acks}

\bibliographystyle{ACM-Reference-Format}
\bibliography{sample}


\begin{thebibliography}{4}


\ifx \showCODEN    \undefined \def \showCODEN     #1{\unskip}     \fi
\ifx \showDOI      \undefined \def \showDOI       #1{#1}\fi
\ifx \showISBNx    \undefined \def \showISBNx     #1{\unskip}     \fi
\ifx \showISBNxiii \undefined \def \showISBNxiii  #1{\unskip}     \fi
\ifx \showISSN     \undefined \def \showISSN      #1{\unskip}     \fi
\ifx \showLCCN     \undefined \def \showLCCN      #1{\unskip}     \fi
\ifx \shownote     \undefined \def \shownote      #1{#1}          \fi
\ifx \showarticletitle \undefined \def \showarticletitle #1{#1}   \fi
\ifx \showURL      \undefined \def \showURL       {\relax}        \fi
\providecommand\bibfield[2]{#2}
\providecommand\bibinfo[2]{#2}
\providecommand\natexlab[1]{#1}
\providecommand\showeprint[2][]{arXiv:#2}

\bibitem[\protect\citeauthoryear{Jing, Huang, Wang, Yao, Yu, Ma, Zhang, Du, and
  Yu}{Jing et~al\mbox{.}}{2025}]%
        {jing2025dsbench}
\bibfield{author}{\bibinfo{person}{Liqiang Jing}, \bibinfo{person}{Zhehui
  Huang}, \bibinfo{person}{Xiaoyang Wang}, \bibinfo{person}{Wenlin Yao},
  \bibinfo{person}{Wenhao Yu}, \bibinfo{person}{Kaixin Ma},
  \bibinfo{person}{Hongming Zhang}, \bibinfo{person}{Xinya Du}, {and}
  \bibinfo{person}{Dong Yu}.} \bibinfo{year}{2025}\natexlab{}.
\newblock \showarticletitle{DSBench: How Far Are Data Science Agents from
  Becoming Data Science Experts?}. In \bibinfo{booktitle}{\emph{ICLR}}.
\newblock


\bibitem[\protect\citeauthoryear{Lai, Vitagliano, Zhang, Sudhir, Chabra, Zeng,
  Zabreyko, Li, Kossmann, Ding, Chen, Markakis, Russo, Wang, Wu, Cafarella,
  Cao, Madden, and Kraska}{Lai et~al\mbox{.}}{2026}]%
        {lai2025kramabench}
\bibfield{author}{\bibinfo{person}{Eugenie Lai}, \bibinfo{person}{Gerardo
  Vitagliano}, \bibinfo{person}{Ziyu Zhang}, \bibinfo{person}{Sivaprasad
  Sudhir}, \bibinfo{person}{Om Chabra}, \bibinfo{person}{Anna Zeng},
  \bibinfo{person}{Anton~A. Zabreyko}, \bibinfo{person}{Chenning Li},
  \bibinfo{person}{Ferdi Kossmann}, \bibinfo{person}{Jialin Ding},
  \bibinfo{person}{Jun Chen}, \bibinfo{person}{Markos Markakis},
  \bibinfo{person}{Matthew Russo}, \bibinfo{person}{Weiyang Wang},
  \bibinfo{person}{Ziniu Wu}, \bibinfo{person}{Michael~J. Cafarella},
  \bibinfo{person}{Lei Cao}, \bibinfo{person}{Samuel Madden}, {and}
  \bibinfo{person}{Tim Kraska}.} \bibinfo{year}{2026}\natexlab{}.
\newblock \showarticletitle{KramaBench: {A} Benchmark for {AI} Systems on
  Data-to-Insight Pipelines over Data Lakes}. In
  \bibinfo{booktitle}{\emph{ICLR}}.
\newblock


\bibitem[\protect\citeauthoryear{Li, Wu, Zhang, Huang, Zhang, Jiang, Wang,
  Zhang, Chen, Shi, Chen, and Li}{Li et~al\mbox{.}}{2025}]%
        {li2025omnisql}
\bibfield{author}{\bibinfo{person}{Haoyang Li}, \bibinfo{person}{Shang Wu},
  \bibinfo{person}{Xiaokang Zhang}, \bibinfo{person}{Xinmei Huang},
  \bibinfo{person}{Jing Zhang}, \bibinfo{person}{Fuxin Jiang},
  \bibinfo{person}{Shuai Wang}, \bibinfo{person}{Tieying Zhang},
  \bibinfo{person}{Jianjun Chen}, \bibinfo{person}{Rui Shi},
  \bibinfo{person}{Hong Chen}, {and} \bibinfo{person}{Cuiping Li}.}
  \bibinfo{year}{2025}\natexlab{}.
\newblock \showarticletitle{OmniSQL: Synthesizing High-quality Text-to-SQL Data
  at Scale}.
\newblock \bibinfo{journal}{\emph{Proc. {VLDB} Endow.}} \bibinfo{volume}{18},
  \bibinfo{number}{11} (\bibinfo{year}{2025}), \bibinfo{pages}{4695--4709}.
\newblock


\bibitem[\protect\citeauthoryear{Lin, Qi, Zhu, Palpanas, Chai, Tang, and
  Luo}{Lin et~al\mbox{.}}{2025}]%
        {lin2025lead}
\bibfield{author}{\bibinfo{person}{Xiaotian Lin}, \bibinfo{person}{Yanlin Qi},
  \bibinfo{person}{Yizhang Zhu}, \bibinfo{person}{Themis Palpanas},
  \bibinfo{person}{Chengliang Chai}, \bibinfo{person}{Nan Tang}, {and}
  \bibinfo{person}{Yuyu Luo}.} \bibinfo{year}{2025}\natexlab{}.
\newblock \showarticletitle{LEAD: Iterative Data Selection for Efficient LLM
  Instruction Tuning}.
\newblock \bibinfo{journal}{\emph{Proc. {VLDB} Endow.}} \bibinfo{volume}{19},
  \bibinfo{number}{3} (\bibinfo{year}{2025}), \bibinfo{pages}{426--439}.
\newblock


\end{thebibliography}

\end{document}